\documentclass{article}

\usepackage{microtype}
\usepackage{graphicx}
\usepackage{booktabs} 
\usepackage{caption}

\usepackage{subcaption}

\usepackage{tabularx}
\usepackage{array}
\newcolumntype{Y}{>{\centering\arraybackslash}X}

\usepackage{hyperref}

\usepackage[preprint]{neurips_2023}

\usepackage{fancyhdr}
\renewcommand{\headrulewidth}{1pt} 
\usepackage{amsmath}
\usepackage{amssymb}
\usepackage{mathtools}
\usepackage{amsthm}

\usepackage{multirow}
\usepackage{booktabs}

\DeclareMathOperator*{\argmax}{arg\,max}

\usepackage{algorithm}
\usepackage{algpseudocode}

\usepackage{xcolor}
\usepackage[most]{tcolorbox}
\tcbuselibrary{skins,breakable}

\definecolor{cfgbg}{RGB}{248,249,251}
\definecolor{cfgframe}{RGB}{214,219,224}

\newtcolorbox{configblock}[1]{
  breakable,
  enhanced,
  colback=cfgbg,
  colframe=cfgframe,
  boxrule=0.6pt,
  arc=2.5mm,
  left=1.2mm,right=1.2mm,top=0.9mm,bottom=0.9mm,
  title=\textbf{#1},
  fonttitle=\small,
}

\definecolor{promptbg}{RGB}{250,247,243}     
\definecolor{promptframe}{RGB}{180,160,140}  

\newtcolorbox{promptblock}[1]{
  breakable,
  enhanced,
  colback=promptbg,
  colframe=promptframe,
  boxrule=0.8pt,
  arc=2.5mm,
  left=1.2mm,right=1.2mm,top=0.9mm,bottom=0.9mm,
  title=\textbf{#1},
  fonttitle=\small,
}

\definecolor{archbg}{RGB}{245,248,252}      
\definecolor{archframe}{RGB}{90,120,160}    
\definecolor{archaccent}{RGB}{70,100,140}   

\newtcolorbox{archblock}[1]{
  breakable,
  enhanced,
  colback=archbg,
  colframe=archframe,
  boxrule=0.9pt,
  arc=2.5mm,
  left=1.4mm,right=1.4mm,top=1.0mm,bottom=1.0mm,
  title=\textbf{#1},
  fonttitle=\small,
}

\usepackage[capitalize,noabbrev]{cleveref}

\theoremstyle{plain}

\theoremstyle{definition}

\theoremstyle{remark}

\usepackage[textsize=tiny]{todonotes}

\newif\ifcomments
\newcommand{\eat}[1]{}
\newcommand{\adaevolve}[1]{{AdaEvolve}}

\commentstrue
\ifcomments
    \providecommand{\shubham}[1]{{\color{red}{/* shubham: #1 */}}}
    \providecommand{\mert}[1]{{\color{olive}{/* mert: #1 */}}}
    \providecommand{\shu}[1]
    {{\color{olive}{/* shu: #1 */}}}
    \providecommand{\accheng}[1]{{\color{blue}{/* accheng: #1 */}}}
    \providecommand{\alex}[1]{{\color{cyan}{/* alex: #1 */}}}
     \providecommand{\ion}[1]{{\color{cyan}{/* ion: #1 */}}}
    \providecommand{\akshat}[1]{{\color{brown}{/* akshat: #1 */}}}
    \providecommand{\eren}[1]{{\color{purple}{/* eren: #1 */}}}
\else
    \providecommand{\shubham}[1]{}
    \providecommand{\mert}[1]{}
    \providecommand{\accheng}[1]{}
    \providecommand{\shu}[1]{}
    \providecommand{\ion}[1]{}
    \providecommand{\alex}[1]{}
    \providecommand{\akshat}[1]{}
    \providecommand{\eren}[1]{}
\fi

\title{\adaevolve{}: Adaptive LLM Driven Zeroth-Order Optimization}

\author{
\textbf{Mert Cemri}$^{1}$\thanks{Equal contribution. Correspondence: cemri@berkeley.edu, shubham3@berkeley.edu}
\quad
\textbf{Shubham Agrawal}$^{1}$\footnotemark[1]
\quad
\textbf{Akshat Gupta}$^{1}$
\quad
\textbf{Shu Liu}$^{1}$ \\
\textbf{Audrey Cheng}$^{1}$
\quad
\textbf{Qiuyang Mang}$^{1}$
\quad
\textbf{Ashwin Naren}$^{1}$
\quad
\textbf{Lutfi Eren Erdogan}$^{1}$ \\
\textbf{Koushik Sen}$^{1}$
\quad
\textbf{Matei Zaharia}$^{1}$
\quad
\textbf{Alex Dimakis}$^{1,2}$
\quad
\textbf{Ion Stoica}$^{1}$ \\
$^{1}$ University of California, Berkeley \\
$^{2}$ Bespoke Labs
}

\begin{document}
\maketitle
\author{}






  
\vspace{-1em}
\begin{abstract}

The paradigm of automated program generation is shifting from one-shot generation to inference-time search, where Large Language Models (LLMs) function as semantic mutation operators within evolutionary loops. While effective, these systems are currently governed by static schedules that fail to account for the non-stationary dynamics of the search process. This rigidity results in substantial computational waste, as resources are indiscriminately allocated to stagnating populations while promising frontiers remain under-exploited.
We introduce AdaEvolve, a framework that reformulates LLM-driven evolution as a hierarchical adaptive optimization problem.
AdaEvolve uses an ``accumulated improvement signal" to unify decisions across three levels: \textbf{Local Adaptation}, which dynamically modulates the exploration intensity within a population of solution candidates; \textbf{Global Adaptation}, which routes the global resource budget via bandit-based scheduling across different solution candidate populations; and \textbf{Meta-Guidance} which generates novel solution tactics based on the previously generated solutions and their corresponding improvements when the progress stalls. 
We demonstrate that AdaEvolve consistently outperforms the open-sourced baselines across 185 different open-ended optimization problems including combinatorial, systems optimization and algorithm design problems \footnote{The code is available at \href{https://github.com/skydiscover-ai/skydiscover.git}{https://github.com/skydiscover-ai/skydiscover.git}}. 

\end{abstract}

\begin{figure*}[h!]
    \centering
    \includegraphics[width=\linewidth]{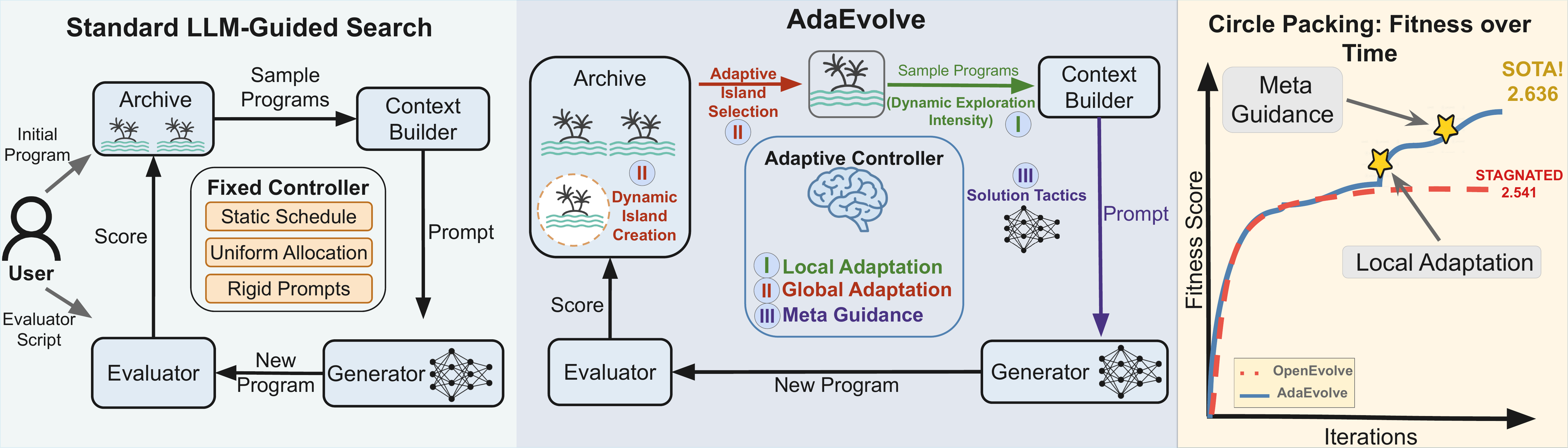}
    \caption{\adaevolve{} overview.
Left: Standard LLM-guided search relies on fixed optimization policies, with static schedules, uniform resource allocation, and rigid prompts.
Center: \adaevolve{} introduces hierarchical adaptivity by dynamically modulating exploration intensity, reallocating compute across populations of programs (islands), and generating meta-level guidance from a unified improvement signal.
Right: \adaevolve{} overcomes stagnation, reaching to a best-known score of 2.636 on the Circle Packing (N = 26), surpassing the Human SOTA (2.634) and AlphaEvolve (2.635).}
    \label{fig:adaevolve_detailed}
\end{figure*}

\section{Introduction}

The frontier of Large Language Model (LLM) research has shifted from scaling training parameters to scaling inference-time compute. It is now established that reasoning performance is not fixed after training but scales with the computation expended during search \citep{snell2024scaling, brown2024large}. This insight has transformed algorithm discovery and program generation from a one-shot generation task into a sequential decision-making problem.


Current non-evolutionary methods for scaling inference compute using a verifier and an LLM based generator such as Beam Search, Monte Carlo Tree Search or Tree-of-Thoughts generally treat the LLM as a static operator, utilizing fixed sampling temperatures and rigid prompt templates throughout
the process. Moreover, these algorithms often struggle to maintain diverse, long-horizon context; a standard Beam Search discards the history of “failed”
attempts, losing valuable information about the search landscape, along with the fact that these test-time scaling methods usually underutilize the LLMs capability to make use of the search semantics by incorporating the feedback
from the past generations when making future generations. 

To alleviate these bottlenecks, LLM-guided Evolutionary
Algorithms (EAs) have emerged as the dominant strategy
for complex program generation, combinatorial optimization and algorithmic discovery \citep{romera2024mathematical, novikov2025alphaevolve}. Unlike traditional EAs with random syntactic mutations, LLMs act as semantic mutation operators to navigate discrete, non-differentiable search spaces.

However, a critical disparity exists in current frameworks. While the mutation operator (the LLM) is sophisticated and context-aware, the  search algorithms controlling it remain surprisingly simple. Canonical systems like OpenEvolve \citep{openevolve} rely on static, pre-determined schedules.
This creates a fragility problem: developers must manually tune hyperparameters like mutation rates, population sizes and solution tactics for every new problem. 
The rigidity of the fixed exploration rates, prompt templates and uniform sampling strategies (determined before the run begins) ignores the non-stationary dynamics of evolutionary search. 
If the parameters are too conservative, the search stagnates in local optima; if too aggressive, it fails to refine solutions. For example, in the Circle Packing benchmark, OpenEvolve fails to converge unless a human operator manually stops the run after 100 iterations and restarts it with a ``refinement" configuration. Because the algorithm cannot adapt its own behavior based on progress, human intuition is required to bridge the gap.

In the domain of continuous optimization, these limitations inspired the idea of adaptive gradient methods like AdaGrad, RMSProp, Adam \citep{duchi2011adaptive, graves2013generating, kingma2014adam}. These algorithms use the first and second moments of the gradients 
to dynamically adjust the learning rates for each parameter, accelerating updates in flat regions and dampening oscillations in steep ones. 
While the LLM-based program generation is a gradient-free (zero-th order) optimization problem, we observe that the trajectory of fitness improvements provides a signal analogous to gradient magnitudes. When a search trajectory yields substantial fitness gains, it signals a productive gradient that should be exploited; when gains vanish, it signals stagnation requiring variance or redirection.


To this end, we introduce \adaevolve{}, a framework that formalizes LLM-driven evolution as a hierarchical dynamic optimization problem. 
A key advantage of AdaEvolve is its minimal configuration requirement: unlike OpenEvolve, which demands per-task or within-run tuning, AdaEvolve requires only the LLM name and iteration count from the user.
Internally, \adaevolve{} maintains an accumulated improvement signal $G_t$ which is updated as an exponential moving average of squared normalized improvements. 
This single signal coordinates adaptation across three coupled levels:
\textbf{Level 1: Local Adaptation} (Within Island Exploration Intensity), \adaevolve{} continuously modulates \emph{Exploration Intensity} ($I_t$) within each subpopulation, automatically shifting from exploration to exploitation as solutions refine, and increasing exploration to as solutions stagnate without manual restart thresholds. At \textbf{Level 2: Global Adaptation} (Across Island Resource Allocation), \adaevolve{} treats computational resources as a dynamic budget. Using a multi-armed bandit with globally normalized rewards, it routes compute power to productive populations of programs (islands) while starving those that have plateaued. Finally, if numerical adaptation is insufficient and global progress stalls, the system triggers a meta-level ``System 2'' intervention, which we call \textbf{Level 3: Meta-Guidance}. At this level, instead of trying to mutate the full code, \adaevolve{} generates new high-level solution tactics to redirect the search toward qualitatively different solution approaches.

\paragraph{Contributions} We make the following contributions to the field of LLM-guided optimization:

\begin{enumerate}
    \item \textbf{An Adaptive Framework for Evolution:} We introduce \adaevolve{}, a novel evolutionary LLM-guided algorithm discovery framework that unifies search intensity, resource allocation and strategy generation under a single adaptive optimizer. By deriving all decisions from the same history based improvement signal, we replace the ad-hoc heuristics of prior works with a cohesive adaptive optimizer.
    \item \textbf{Dynamic, Globally-Normalized Resource Allocation.} We propose a new bandit-based scheduler that routes computational budget to the most productive subpopulations. Crucially, we introduce a \emph{global normalization} mechanism that evaluates improvements relative to the global best solution rather than local history. This prevents resources from being wasted on islands that are merely refining poor solutions (local optima) or resting on the successes of old stale generations, ensuring compute is always directed toward the current frontier of the search.
    \item \textbf{Robust and Strong Generalization.} \adaevolve{} improves over baselines across 185 different optimization/algorithm discovery problems spanning combinatorial geometry, systems optimization, and algorithm design, using identical hyperparameters throughout. 
    In particular, \adaevolve{} reaches or matches the best known human or prior AI solutions (including proprietary models like AlphaEvolve) in 4/6 mathematical optimization tasks, achieves human-competitive or superior performance in 6/7 ADRS systems benchmarks, while consistently improving over the open source baselines in all problems. 
\end{enumerate}

\section{Related Works}
\label{sec:background}

\paragraph{Test-Time Scaling and Search Algorithms.}
Test-time compute scaling laws suggest that increased inference budget can improve model performance \citep{snell2024scaling}.
Techniques such as Chain-of-Thought \citep{wei2022chain} and Self-Consistency \citep{wang2022self} exploit this by
sampling diverse reasoning paths, while more structured approaches like Monte Carlo Tree Search (MCTS)
\citep{zhang2024rest, chopra2025feedback} explicitly build search trees to navigate the solution space. Works such as
\cite{abe2025llm, li2024agent, liang2024encouraging, du2023improving} studied multi-agent scaffolds around LLMs to
facilitate complex reasoning, including inference-time evolutionary reasoning mechanisms \citep{lee2025evolving}.

\paragraph{LLM-Guided Evolutionary Search.}
For automated discovery, LLMs have been incorporated at inference time with search-and-feedback scaffolds using
evolutionary optimization that iteratively propose, evaluate, and refine candidate solutions, defining the modern era of
Genetic Programming \citep{langdon2013foundations, koza1994genetic}. FunSearch \citep{romera2024mathematical} and
Evolution through Large Models (ELM) \citep{lehman2023evolution} demonstrated that LLMs could act as semantic
variation operators by solving open combinatorial problems. Subsequently, AlphaEvolve
\citep{novikov2025alphaevolve} generalized this approach by maintaining populations of candidate programs iteratively
improved via mutation, crossover, and selection across scientific and engineering problems.

GEPA \citep{agrawal2025gepa} is closely related in spirit to AlphaEvolve-style scaffolds, but targets optimizing prompts
of compound LLM systems via LLM-based reflection while maintaining a Pareto set of strong-but-diverse candidates.
Related directions also explore evolutionary refinement directly in language space
\citep{guo2023evoprompt}, including mechanisms that evolve mutation operators or generation policies
\citep{fernando2023promptbreeder}. Reflection-guided evolutionary dynamics have also been studied
\citep{ye2024reevo}.

Recent work has introduced open-source island-based evolutionary frameworks, including OpenEvolve
\citep{openevolve} and ShinkaEvolve \citep{lange2025shinkaevolve}. While OpenEvolve follows AlphaEvolve
closely, ShinkaEvolve targets sample efficiency through improved parent sampling, rejection-sampled code rewrites,
and adaptive LLM ensembles. CodeEvolve \citep{assumpccao2025codeevolve} similarly studies LLM-guided program
evolution within an island-based genetic algorithm framework. SOAR~\citep{pourcel2025self} studies self-improving
evolutionary program synthesis via hindsight fine-tuning, while DeltaEvolve~\citep{jiang2026deltaevolve} explores
context-efficient evolutionary updates. Complementary work studies progress-aware evolution \citep{yan2026pacevolve}.
However, these existing methods rely on largely fixed resource allocation decided before evolution. \adaevolve{} instead introduces an adaptive evolutionary framework that dynamically regulates the
exploration-exploitation tradeoff and mitigates progress stagnation without manual threshold tuning.
Variants such as ThetaEvolve \citep{wang2025thetaevolve}, FLEX \citep{cai2025flex} and TTT-Discover
\citep{yuksekgonul2026learning}
explore learning-based adaptations, whereas our work focuses on inference-only evolutionary scaffolds.

\paragraph{Adaptive Optimization and Control.}
Adaptive Gradient Methods in continuous optimization, such as Adam \citep{kingma2014adam}, RMSProp
\citep{ruder2016overview}, and AdaGrad \citep{duchi2011adaptive}, utilize exponential moving averages of gradient
moments to normalize updates. \adaevolve{} applies a similar principle to discrete search by using the trajectory of
fitness improvements as a gradient analogue. \cite{chen2023symbolic} used a tournament-based evolutionary algorithm
with syntactic mutations to discover optimization algorithms, finding the Lion optimizer. Adaptive Operator Selection
(AOS) \citep{fialho2010adaptive} approaches traditionally assign credit to mutation operators based on recent
performance. Related work also studies structured context evolution and adaptation mechanisms \citep{zhang2025agentic, suzgun2025dynamic}. In contrast, \adaevolve{} proposes a unified adaptive paradigm for the open-ended semantic space of LLMs: rather than
selecting between fixed operators, our system uses a unified improvement signal to dynamically modulate search
intensity, global resource budget, and meta-level guidance. This formulation elevates adaptation to higher levels of the
optimization hierarchy, allowing the system to autonomously regulate exploration-exploitation dynamics and replace
brittle static schedules of prior frameworks.

\section{The \adaevolve{} Framework}
\label{sec:method}


\subsection{Problem Formulation}

We formalize LLM-driven program synthesis as a hierarchical optimization problem. The objective is to maximize a fitness function $\mathcal{F}: \mathcal{P} \to \mathbb{R}$ over a discrete space of executable programs $\mathcal{P}$, subject to a computational budget $B$ (which may be measured in iterations or LLM calls).

The search is distributed across a dynamic set of $K$ parallel subpopulations, which we call \textbf{islands}. Each island operates asynchronously, maintaining its own local archive of programs, denoted as $\mathcal{D} = \{D_1, \dots, D_K\}$. At any time step $t$, an island $k$ executes an evolutionary cycle:

\textbf{1) Selection:} Sample a parent program $p$ from the island's archive $D_k$\\
\textbf{2) Mutation:} Use an LLM to generate a child program $p'$ \\
\textbf{3) Evaluation:} Compute the child's fitness $f' = \mathcal{F}(p')$ \\
\textbf{4) Update:} Add $p'$ to the archive $D_k$ and update the adaptive state \\

\adaevolve{} controls this iterative process. A critical distinction of \adaevolve{} is the elimination of manual configuration. Unlike prior frameworks that require users to tune complex configuration files prior to a run (e.g., exploration rates, island counts, prompt templates), \adaevolve{} treats these as internal dynamic variables. The only inputs required from the user apart from the evaluator and problem specification are the model name and the total iteration budget. \adaevolve{} modulates the search dynamics through three synchronized feedback loops: Level 1 Local Adaptation (within-island exploration intensity), Level 2 Global Adaptation (across-island resource allocation), and Level 3 Meta-Guidance for generating new solution tactics.

\subsection{Level 1: Local Adaptation (Within Island Exploration Intensity)}

Within each island, the fundamental challenge is balancing \emph{exploration} (searching new regions of the fitness landscape) with \emph{exploitation} (refining promising solutions). Rather than using a fixed ratio, \adaevolve{} continuously adapts this value based on the island's recent productivity.

\paragraph{The Accumulated Improvement Signal.}
For each island $k$, we track a scalar signal $G_t^{(k)}$ that summarizes the island's recent improvement history. When a new program $p'$ is generated by an LLM mutation operator, we evaluate its fitness as $f'$ and compute a normalized improvement magnitude:
\begin{align}
    \delta_t^{(k)} = \max\left(\frac{f' - f^*_k}{f^*_k} , 0 \right)
\end{align}
Here, $f^*_k$ denotes the current best of island $k$. This normalization makes the signal scale-invariant, helping the algorithm generalize across different use-cases.

The accumulated improvement signal is then updated as an exponential moving average of squared improvements:
\begin{align}
G_t^{(k)} = \rho \cdot G_{t-1}^{(k)} + (1 - \rho) \cdot (\delta_t^{(k)})^2
\label{eq:G_update}
\end{align}
where $\rho \in [0, 1)$ is a decay factor. During periods of stagnation, (where $f' \leq f^*_k$),  $\delta_t = 0$, so $G_t^{(k)} = \rho G_{t-1}^{(k)}$, and the signal decays exponentially. Thus, the accumulated improvement signal $G_t^{(k)}$ acts as a real-time volatility metric: high values indicate a productive trajectory (``steep gradient"), while low values indicate convergence or stagnation.

\paragraph{Dynamic Exploration Intensity.}
We use $G_t^{(k)}$ to compute the \emph{exploration intensity} $I_t^{(k)} \in [I_{\min}, I_{\max}]$ that acts as the probability of exploration:
\begin{align}
I_t^{(k)} = I_{\min} + \frac{I_{\max} - I_{\min}}{1 + \sqrt{G_t^{(k)} + \epsilon}}
\label{eq:intensity}
\end{align}
where $I_{\min}$ and $I_{\max}$ are hyperparameters defining the range of exploration probabilities (we use $I_{\min} = 0.1$ and $I_{\max} = 0.7$).
High $G_t^{(k)}$ indicates that the island $k$ is productive at iteration $t$, so we have $I_t^{(k)} \to I_{\min}$, favoring exploitation of the current productive trajectory. Conversely, low  $G_t^{(k)}$ indicates stagnation, so we have $I_t \to I_{\max}$, increasing exploration to escape local optima.

At each iteration, we sample the search mode stochastically: with probability $I_t$, the algorithm does \emph{exploration} and with probability $1 - I_t$ the algorithm does \emph{exploitation}. During exploration, the parents are selected uniformly at random from the islands archive, and we use an exploration prompt that encourages more orthogonal solutions compared to the sampled programs from the island's archive. During exploitation, parents are selected with proportion to their fitness values, and the mutation operator is prompted to do refinements on the sampled programs.

\subsection{Level 2: Global Adaptation (Across Island Resource Allocation)}

While Level 1 optimizes \emph{how} each island searches, Level 2 optimizes \emph{where} the global computational budget is allocated. We model this as a multi-armed bandit problem where each island is an ``arm", and the goal is to route compute to the islands most likely to yield future improvements.

\paragraph{Decayed-Magnitude Bandit-Based Island Selection.}
A naive bandit approach using each island's local improvement ($\delta_t^{(k)}$) as the reward signal fails as it biases selection toward islands with low baseline fitness making trivial refinements (this can also be called as ``poor island bias").
Consider the following example: Island 1 at fitness 100 finds a +10 improvement, yielding $\delta^{(1)} = 0.10$. Island 2 at fitness 1 finds a +0.5 improvement, yielding $\delta^{(2)} = 0.50$. A bandit using these rewards would favor Island 2 despite Island 1's improvement being more valuable globally.

To resolve this, \adaevolve{} normalizes bandit rewards by the \emph{global} best fitness $f^*_{\text{global}}$ (the best across all islands) rather than each island's local best. When island $k$ improves from $f_k^*$ to $f'$, the bandit reward $r_t^{(k)}$ for an improvement on island $k$ is defined as:
\begin{align}
r_t^{(k)} = \frac{f' - f_k^*}{f^*_{\text{global}}}
\label{eq:global_reward}
\end{align}
This ensures resources are directed toward globally significant progress; a unit of improvement is valued equally regardless of which island produces it.


To prevent islands that are currently stale but had early breakthroughs from dominating allocation decisions, we maintain {decayed} cumulative rewards $R^{(k)}$ and decayed visit counts $V^{(k)}$ for each island, updated at each iteration as:
\begin{align}
R_t^{(k)} &= \rho \cdot R_{t-1}^{(k)} + r_t^{(k)}, \quad
V_t^{(k)} = \rho \cdot V_{t-1}^{(k)} + 1
\label{eq:decay_stats}
\end{align}
We select islands using an Upper Confidence Bound (UCB). Let $n_k$ denote the visit count for island $k$, and $N = \sum_k n_k$ the total iterations. The selection rule is:
\begin{align}
k^* = \argmax_k \left[ \frac{R_k}{V_k} + C \sqrt{\frac{\ln N}{n_k}} \right]
\label{eq:ucb}
\end{align}
where $C = \sqrt{2}$ is the exploration constant for the island selection. The ratio $R_k / V_k$ reflects recent productivity rather than lifetime productivity. Any island with $V^{(k)} = 0$ must be visited at least once. If multiple unvisited islands exist, they are selected uniformly at random.

\paragraph{Migration Across Islands.}

Every $M$ iterations, islands exchange their top programs via ring migration: island $k$ sends its best programs to island $\left ( (k+1) \mod K \right)$. Migrated programs update the receiving island's local best $f_k^*$ and accumulated signal $G^{(k)}$ (ensuring correct intensity adaptation), but they do \emph{not} update UCB statistics since the receiving island did not generate the improvement.

\paragraph{Dynamic Island Spawning.}
\adaevolve{} dynamically creates new islands when the productivity across all islands stagnates. In particular, when $G_t^{(k)} \leq \tau_S$ for all islands $k$ at an iteration $t$ for some island spawning threshold $\tau_S$, we decleare all islands stagnated and spawn a new one. In our experiments, we always set $\tau_S=0.02$ for all experiments and show that it is able to get us SOTA results in almost all 180+ problems. In general, islands maintain diversity through isolation and infrequent migration. However, unlike other frameworks that typically use a fixed number of islands, {we show that creating new subpopulations adaptively based on need of stagnation improves resource allocation}. 
\adaevolve{} uses the current islands until improvement stops and then initializes a new island with a randomly sampled seed program from the archive to explore alternative solutions.


\subsection{Level 3: Meta-Guidance (Solution Tactics Generation)}
If numerical adaptation (Level 1) and resource reallocation (Level 2) fail to yield progress, it implies the search is trapped in a conceptual local optimum: the code is optimized, but the underlying algorithm is suboptimal. Level 3 addresses this by generating solution tactics: high-level algorithmic directives that force a qualitative shift in the search trajectory. When global stagnation is detected, AdaEvolve triggers a ``System 2" meta-analysis. 
This Meta-Guidance for generating solution tactics get triggered when $G_t^{(k)} \leq \tau_M$ for all islands $k$ at an iteration $t$ for some island Meta-Guidance threshold $\tau_M$. In our experiments, we always set $\tau_M=0.12$ for all experiments.

In this level of adaptation, 
\adaevolve{} invokes a separate LLM to analyze the problem specification, evaluator, and recent failed attempts. This meta-analysis identifies \emph{why} current approaches are insufficient and proposes alternative strategies (e.g., ``switch from greedy selection to dynamic programming''). These tactics are then injected into mutation prompts, transforming the task from open-ended improvement to targeted implementation of a specific strategy. If a solution tactic fails to yield progress, the system rotates to alternatives before eventually generating new tactics. \textbf{This provides a mechanism for escaping solution-space bottlenecks that numerical adaptation alone cannot resolve}.
We examples of different Meta-Guidance generations in \cref{tab:paradigm-examples}.

\subsection{Algorithm Summary}
Algorithm~\ref{alg:adaevolve} presents the complete procedure. The three levels operate at different timescales: intensity adapts every iteration, island selection operates across iterations, and Meta-Guidance generation triggers when numerical adaptation alone proves insufficient. The additional subroutines in Algorithm \ref{alg:adaevolve} are explained in \Cref{app:subroutines}.

For island $k$ at iteration $t$: $G_t^{(k)}$ is the accumulated improvement signal, $f_k^*$ is the local best fitness, $f^*_{\text{global}}$ is the best fitness score across all islands, $R_t^{(k)}$ and $V_t^{(k)}$ are the decayed reward and visit counts, and $n_k$ is the raw visit count.

\begin{algorithm}[t]
\caption{\adaevolve{} -- Main Loop}
\label{alg:adaevolve}
\begin{algorithmic}[1]

\Require Initial program $p_0$, evaluator $\mathcal{F}$, budget $T$, starting num of islands $K$, LLM $M$

\State \textbf{Initialize} $\forall k \in \{1,\ldots,K\}$:
\State \quad $G_0^{(k)} \gets 0$, $R_0^{(k)} \gets 0$, $V_0^{(k)} \gets 0$, $n_k \gets 0$
\State \quad $f_k^* \gets \mathcal{F}(p_0)$, $D_k \gets \{p_0\}$
\State $f^*_{\text{global}} \gets \mathcal{F}(p_0)$, $\mathcal{T}=\emptyset$

\For{$t = 1$ to $T$}

    \Statex \textcolor{blue}{\texttt{// Level 2: Select island via UCB}}

    \State $k \gets \textsc{SelectIsland}(t)$ \hfill $\triangleright$ Eq.~\ref{eq:ucb}

    \Statex \textcolor{blue}{\texttt{// Level 1: Compute intensity}}
    
    \State $I_t^{(k)} \gets I_{\min} + \frac{I_{\max} - I_{\min}}{1 + \sqrt{G_{t-1}^{(k)} + \epsilon}}$
    \hfill $\triangleright$ Eq.~\ref{eq:intensity}

    \Statex \textcolor{blue}{\texttt{// Sample and mutate}}

    \State $p$, inspires $\gets$ \textsc{Sample}(${D}_k$, $I_t^{(k)}$)
    \hfill $\triangleright$ Alg.~\ref{alg:sample}

    \State prompt $\gets$ \textsc{ContextBuilder}($p$, inspires, $\mathcal{T}$)
    \hfill $\triangleright$ Alg.~\ref{alg:context_builder}
    
    \State $p'$ $\gets$ \textsc{Mutate}($M$, prompt)
    \hfill $\triangleright$ Alg.~\ref{alg:mutate}

    \State $f' \gets \mathcal{F}(p')$

    \State $D_k \gets D_k \cup \{p'\}$

    \Statex \textcolor{blue}{\texttt{// Update adaptive state}}

    \State \textsc{UpdateState}($k$, $f'$)
    \hfill $\triangleright$ Alg.~\ref{alg:update}

    \State \textcolor{blue}{\texttt{// Level 3: Meta-control}}
    \If{\textsc{GloballyStagnant}() $\land$ paradigm $=$ \textsc{None}}
        \State $\mathcal{T}$ $\gets$ \textsc{GenerateParadigm}() \hfill $\triangleright$ Alg.~\ref{alg:paradigm}
    \EndIf

\EndFor

\State \Return $\argmax_{p} \mathcal{F}(p)$ over all archives

\end{algorithmic}
\end{algorithm}


\newpage

\section{Experiments}
\label{sec:experiments}

We evaluate \adaevolve{} on 185 different algorithm design/ optimization problems. These problems are from: mathematical optimization problems (6) discussed in \Cref{sec:math_benchmarks}; real-world system optimization problems (7) from the ADRS benchmark suite discussed in \Cref{sec:adrs_benchmarks}; open-ended challenging algorithm design problems (172) from Frontier-CS benchmark suite discussed in \Cref{sec:frontier_cs}. We also report numbers on the ARC-AGI-2 task in \Cref{sec:arc_agi}.
In all these benchmarks, we demonstrate that \adaevolve{} outperforms strong baselines over diverse benchmarks. 
In \Cref{sec:ablations}, we provide ablation studies demonstrating the contributions of different components of the \adaevolve{} algorithm via comprehensive analysis on two benchmarks. Finally, in \Cref{sec:case_studies}, we provide case studies of different regimes where the combined adaptivity levels and Meta-Guidance yield the largest gains.


\subsection{Experimental Setup}

For the mathematical optimization/algorithm development experiments in \Cref{sec:math_benchmarks} and ADRS benchmark in \Cref{sec:adrs_benchmarks},
all methods are evaluated using the same backbone model GPT-5 and Gemini-3-Pro, the same evaluator for the fitness scoring,
and total iteration budget of $T=100$ iterations. We report the mean $\pm$ standard deviation over three independent runs as well as the best of these three runs.
Because of the great number of problems in Frontier-CS benchmark discussed in \Cref{sec:frontier_cs}, we use GPT-5 for all the methods across the 172 problems in that benchmark, and run each model for the same number of 50 LLM calls.

We compare \adaevolve{} against the SOTA evolutionary algorithms with open-source codebases, which are GEPA, ShinkaEvolve and OpenEvolve (introduced in \Cref{sec:background}). Whenever applicable, we also report the results of AlphaEvolve and the best human-made results on these benchmarks.

\subsection{Mathematical Optimization Benchmarks}
\label{sec:math_benchmarks}

Table~\ref{tab:math-opt} reports results on six mathematical combinatorial optimization problems. 
These tasks span a range of optimization regimes,
from relatively smooth objectives to highly deceptive landscapes with many poor local optima. 
A description of these problems are provided in \Cref{tab:adaevolve_problem_defs}. 

\begin{figure*}[h!]
    \centering
    \includegraphics[width=\linewidth]{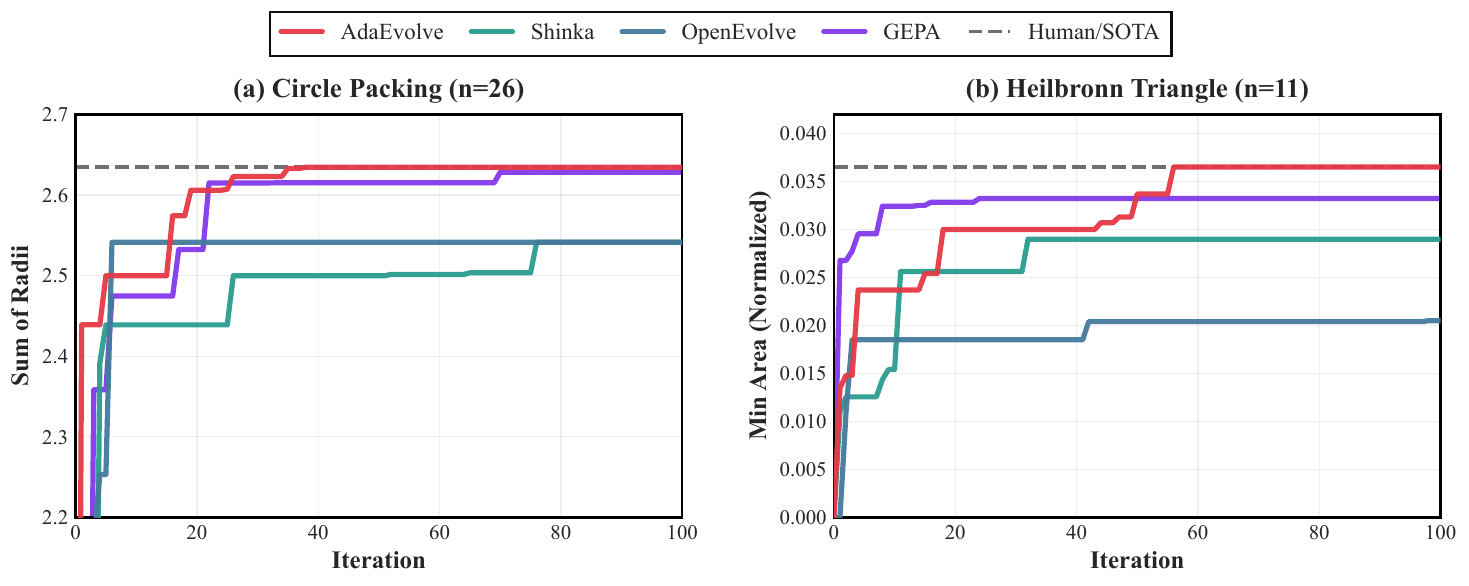}
    \caption{Comparison of the evolutionary algorithms on Circle Packing (Square $n=26$) and Heilbronn Triangles ($n=11$) problems using GPT-5 backbone for all of them. $n$ is a parameter of the optimization problems we explain in \cref{tab:adaevolve_problem_defs}.}
    \label{fig:progress_comparison}
\end{figure*}

\begin{table*}[t]
\centering
\scriptsize 
\setlength{\tabcolsep}{2.5pt} 
\renewcommand{\arraystretch}{1.1} 
\caption{Mathematical optimization benchmarks. We report \textbf{Mean $\pm$ Std} and \textbf{Best} (Max) objective values. Higher is better for all metrics. \textbf{Bold} denotes the best performing method per backbone. Underline denotes results surpassing \textbf{Human SOTA}. ($N$ values: Square=26, Rect=21, Tri=11, Convex=13, Max=3). The detailed explanations are given in \cref{tab:adaevolve_problem_defs}. For circle packing (rect) problem with Gemini, \adaevolve{} gets 2.36583237, beating the AlphaEvolve reference of 2.36583213. For circle packing (square) problem with Gemini, \adaevolve{} gets 2.63598308, beating the AlphaEvolve reference of 2.63586276.}
\label{tab:math-opt-final}

\makebox[\textwidth][c]{\begin{tabular}{l cc cc cc cc cc cc}
\toprule
& \multicolumn{2}{c}{\textbf{Circle Packing (Square)}} 
& \multicolumn{2}{c}{\textbf{Circle Packing (Rect)}} 
& \multicolumn{2}{c}{\textbf{Heilbronn (Triangles)}} 
& \multicolumn{2}{c}{\textbf{Heilbronn (Convex)}} 
& \multicolumn{2}{c}{\textbf{MinMax Distance}} 
& \multicolumn{2}{c}{\textbf{Signal Processing}} \\
\cmidrule(lr){2-3}
\cmidrule(lr){4-5}
\cmidrule(lr){6-7}
\cmidrule(lr){8-9}
\cmidrule(lr){10-11}
\cmidrule(lr){12-13}

\textbf{Strategy} 
& Avg & Best 
& Avg & Best 
& Avg & Best 
& Avg & Best 
& Avg & Best 
& Avg & Best \\
\midrule
\textbf{Human} & -- & 2.634 & -- & 2.364 & -- & 0.0360 & -- & 0.0306 & -- & 0.2399 & -- & --  \\
\textbf{AlphaEvolve} & -- & 2.635 & -- & 2.3658 & -- & 0.0365 & -- & 0.0309 & -- & 0.2398 & -- & --  \\
\midrule
\multicolumn{13}{l}{\textit{Backbone: GPT-5}} \\
OpenEvolve & 2.531 $\pm$ .018 & 2.541 & 2.267 $\pm$ .014 & 2.276 & 0.028 $\pm$ .006 & 0.028 & 0.025 $\pm$ .005 & 0.027 & 0.226 $\pm$ .003 & 0.2243 & 0.569 $\pm$ .047 & 0.622 \\
GEPA & 2.613 $\pm$ .022 & 2.628 & 2.326 $\pm$ .023 & 2.354 & 0.031 $\pm$ .002 & 0.032 & 0.025 $\pm$ .002 & 0.027 & 0.232 $\pm$ .120 & 0.2392 & 0.689 $\pm$ .014 & 0.705 \\
ShinkaEvolve & 2.464 $\pm$ .083 & 2.541 & 2.335 $\pm$ .026 & 2.358 & 0.032 $\pm$ .012 & {0.034} & 0.023 $\pm$ .005 & 0.026 & 0.239 $\pm$ .001 & \textbf{0.2398} & 0.485 $\pm$ .044 & 0.533 \\
\adaevolve{} & \textbf{2.629} $\pm  $ .001 & \textbf{\underline{2.636}} & \textbf{2.349} $\pm$ .010 & \textbf{2.361} & \textbf{0.033} $\pm$ .001 & \textbf{0.036} & \textbf{0.027} $\pm$ .001 & \textbf{0.029} & \textbf{0.239} $\pm$ .009 & \textbf{0.2404} & \textbf{0.698} $\pm$ .017 & \textbf{0.718} \\
\midrule
\multicolumn{13}{l}{\textit{Backbone: Gemini-3-Pro}} \\
OpenEvolve & 2.541 $\pm$ .000 & 2.541 & 2.365 $\pm$ .001 & 2.366 & 0.033 $\pm$ .000 & 0.035 & 0.028 $\pm$ .009 & 0.028 & 0.240 $\pm$ .000 & 0.2398 & 0.553 $\pm$ .015 & 0.565 \\
GEPA & 2.620 $\pm$ .002 & 2.621 & 2.216 $\pm$ .046 & 2.232 & 0.031 $\pm$ .002 & 0.033 & 0.023 $\pm$ .001 & 0.027 & 0.213 $\pm$ .008 & 0.2178 & \textbf{0.617} $\pm$ .075 & \textbf{0.680} \\
ShinkaEvolve & 2.622 $\pm$ .012 & 2.636 & 2.366 $\pm$ .000 & 2.366 & 0.035 $\pm$ .001 & 0.036 & 0.028 $\pm$ .001 & \textbf{0.029} & \textbf{0.240} $\pm$ .000 & {0.2398} & 0.480 $\pm$ .039 & 0.505 \\
\adaevolve{} & \textbf{\underline{2.632}} $\pm$ .003 & \textbf{\underline{2.636}} & \textbf{{2.366}} $\pm$ .000 & \textbf{\underline{2.366}} & \textbf{0.036} $\pm$ .001 & \textbf{0.036} & \textbf{0.029} $\pm$ .000 & \textbf{0.029} & \textbf{0.240} $\pm$ .000 & \textbf{\underline{0.2404}} & 0.593 $\pm$ .009 & 0.603 \\
\bottomrule
\end{tabular}
}
\label{tab:math-opt}
\end{table*}

\paragraph{Results.}
Across all problems in \Cref{tab:math-opt}, \adaevolve{} achieves the best results over other open-source baselines and sometimes exceeds the results of the Human/AlphaEvolve solutions.

\begin{figure}[h]
    \centering
    \includegraphics[width=\linewidth]{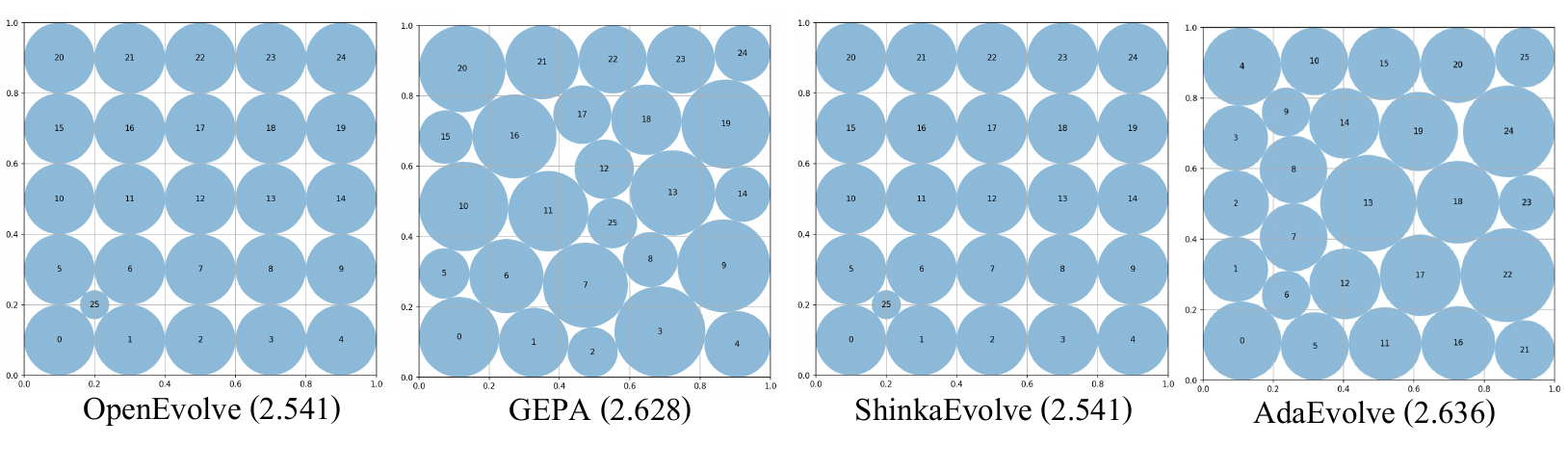}
    \caption{GPT-5 best configurations comparisons for the \texttt{Circle Packing} experiment.}
    \label{fig:circle_packing_illustration}
\end{figure}

The largest gains appear on problems with deceptive fitness landscapes,
such as \texttt{Heilbronn Triangle}
and \texttt{MinMaxDist},
where fixed-policy baselines (OpenEvolve, GEPA, and ShinkaEvolve)
frequently plateau after early progress.
We show the comparison of these methods in \cref{fig:progress_comparison} and an illustration of the best programs for the \texttt{Circle Packing (square)} problem generated by GPT-5 for all the open-source methods in Figure \ref{fig:circle_packing_illustration}. Note that we ran longer (150 total LLM calls), GEPA also reaches 2.63598 in \texttt{Circle Packing (square)}.

\subsection{ADRS Systems Benchmarks}
\label{sec:adrs_benchmarks}

\begin{table*}[t]
\centering
\scriptsize
\setlength{\tabcolsep}{1.8pt}
\renewcommand{\arraystretch}{1.1} 
\caption{Comparison on Systems \& Data benchmarks. We report \textbf{Mean $\pm$ Std} and \textbf{Best}. Higher is better for all metrics except \textbf{Cloudcast} ($\downarrow$). \textbf{Bold} indicates best automated strategy. \underline{Underline} indicates results surpassing \textbf{Human SOTA}. Detailed explanations of these benchmarks are given in \cref{tab:adrs_systems}.}
\label{tab:systems-bench}

\makebox[\textwidth][c]{\begin{tabular}{l cc cc cc cc cc cc cc}
\toprule
& \multicolumn{2}{c}{\textbf{Telemetry}} & \multicolumn{2}{c}{\textbf{Cloudcast} $\downarrow$} & \multicolumn{2}{c}{\textbf{EPLB}} & \multicolumn{2}{c}{\textbf{Prism}} & \multicolumn{2}{c}{\textbf{LLM-SQL}} & \multicolumn{2}{c}{\textbf{TXN}} & \multicolumn{2}{c}{\textbf{NS3}} \\
\cmidrule(lr){2-3}\cmidrule(lr){4-5}\cmidrule(lr){6-7} \cmidrule(lr){8-9}\cmidrule(lr){10-11}\cmidrule(lr){12-13}\cmidrule(lr){14-15}
\textbf{Strategy} & Avg & Best & Avg & Best & Avg & Best & Avg & Best & Avg & Best & Avg & Best & Avg & Best \\
\midrule
\textbf{Human / SOTA} & --  & 0.822 & -- & 626.2 & -- & 0.1265 & -- & 21.89 & -- & 0.692 & -- & 2,725 & -- & 69.0 \\
\midrule
\multicolumn{15}{l}{\textit{Backbone: GPT-5}} \\
OE & .930 $\pm$ .04 & .952 & 926.9 $\pm$ 171 & 729.8 & .127 $\pm$ .00 & .1272 & 26.23 $\pm$ .00 & 26.23 & .710 $\pm$ .01 & .716 & 4,239 $\pm$ 90 & 4,329 & 92.2 $\pm$ 4.8 & 97.3 \\
GEPA & .916 $\pm$ .05 & .948 & 689.9 $\pm$ 74 & 645.7 & .134 $\pm$ .01 & .1445 & 26.19 $\pm$ .07 & 26.23 & .713 $\pm$ .00 & .713 & 3,753 $\pm$ 204 & 3,984 & 68.9 $\pm$ .00 & 101.8 \\
Shinka & .923 $\pm$ .04 & .952 & 954.8 $\pm$ 125 & 812.7 & .118 $\pm$ .01 & .1272 & 26.26 $\pm$ .00 & 26.26 & .712 $\pm$ .00 & .713 & 4,090 $\pm$ 338 & 4,329 & 89.5 $\pm$ 18.7 & 106.1 \\
\adaevolve{} & \textbf{\underline{.952}} $\pm$ .00 & \textbf{\underline{.952}} & \textbf{662.3} $\pm$ 0.1 & \textbf{640.5} & \textbf{\underline{.134}} $\pm$ .01 & \textbf{\underline{.1453}} & \textbf{\underline{26.30}} $\pm$ .07 & \textbf{\underline{26.37}} & \textbf{\underline{.746}} $\pm$ .03 & \textbf{\underline{.775}} & \textbf{\underline{4,317}} $\pm$ 29 & \textbf{\underline{4,348}} & \textbf{\underline{125.2}} $\pm$ 6.1 & \textbf{\underline{131.8}} \\
\midrule
\multicolumn{15}{l}{\textit{Backbone: Gemini-3-Pro}} \\
OE & \textbf{\underline{.954}} $\pm$ .01 & \textbf{\underline{.960}} & 707.8 $\pm$ 40 & 667.1 & .127 $\pm$ .00 & .1272 & 26.24 $\pm$ .01 & 26.24 & .729 $\pm$ .01 & .736 & 4,109 $\pm$ 254 & 4,274 & 115.2 $\pm$ 13.2 & 125.6 \\
GEPA & .850 $\pm$ .00 & .855 & 720.4 $\pm$ 46 & 667.1 & .127 $\pm$ .00 & .1272 & 26.16 $\pm$ .03 & 26.19 & .713 $\pm$ .00 & .713 & 3,616 $\pm$ 481 & 4,167 & 74.5 $\pm$ 9.5 & 104.0 \\
Shinka & .918 $\pm$ .03 & .933 & 949.8 $\pm$ 73 & 892.3 & .120 $\pm$ .01 & .1272 & 26.25 $\pm$ .01 & 26.26 & .721 $\pm$ .00 & .721 & 3,932 $\pm$ 343 & 4,255 & 84.7 $\pm$ 7.7 & 92.2 \\
\adaevolve{} & \underline{.953} $\pm$ .01 & \textbf{\underline{.960}} & \textbf{642.1} $\pm$ 5.9 & \textbf{637.1} & \textbf{\underline{.145}} $\pm$ .00 & \textbf{\underline{.1453}} & \textbf{\underline{26.26}} $\pm$ .00 & \textbf{\underline{26.26}} & \textbf{\underline{.743}} $\pm$ .01 & \textbf{\underline{.752}} & \textbf{\underline{4,221}} $\pm$ 89 & \textbf{\underline{4,310}} & \textbf{\underline{126.0}} $\pm$ 5.1 & \textbf{\underline{131.8}} \\
\bottomrule
\end{tabular}
}
\end{table*}

Table~\ref{tab:systems-bench} reports results on seven real-world systems optimization tasks
from the ADRS \cite{cheng2025barbarians} benchmark suite. 
ADRS includes various tasks such as developing MoE expert balancing (Expert Parallelism Load Balancing - EPLB) problems, transaction scheduling problems or algorithm design for minimizing the cost of multi-cloud data storage problems.
These tasks involve expensive evaluations, noisy feedback,
and heterogeneous objective scales, 
making fixed exploration schedules and static resource allocation brittle.

\paragraph{Results.}
\adaevolve{} achieves the strongest aggregate performance,
winning on all seven tasks across both GPT-5 and Gemini-3-Pro model backbones.
The largest gains occur on tasks characterized by sparse or bursty improvements, such as \texttt{TXN}, \texttt{CBL}, and \texttt{CBL-Multi}, where fixed strategies either over-exploit early trajectories or fail to reallocate resources after prolonged stagnation. For example, on \texttt{TXN} with GPT-5,
\adaevolve{} improves performance from 4329 to 4348,
while fixed baselines plateau early.

On tasks with smoother reward signals,
including \texttt{Prism} and \texttt{LLM-SQL},
performance differences are smaller,
and \adaevolve{} matches or slightly improves upon the strongest fixed baselines.
This indicates that adaptivity does not harm performance
when static strategies are already well-aligned with the task.

Importantly, performance trends are consistent across backbones:
relative rankings among methods are largely preserved
between GPT-5 and Gemini-3-Pro,
suggesting that the gains from adaptive control
do not depend on backbone-specific behavior.

These results show that adaptive optimization is particularly effective
for real-world systems tasks with tight evaluation budgets
and infrequent high-magnitude improvements.




\newpage

\subsection{Frontier-CS}
\label{sec:frontier_cs}

We also evaluate our framework on Frontier-CS \cite{mang2025frontiercs}, a challenging benchmark comprised of 172 open-ended computer science problems, ranging from algorithmic optimization to research-level systems tasks, where the global optima are unknown and solutions are evaluated via executable programs rather than static outputs. We used the same GPT-5 model in all methods with equal number (50) of LLM calls. We also include the result of GPT-5 (single-model) model to demonstrate the remarkable capability of these scaffolds, with \adaevolve{} increasing mean performance by 3x. Note that due to the hardness of these problems, GPT-5 with single calls gets Median score of 0, meaning more than half of the solutions score 0, showing the necessity of such search algorithms like \adaevolve{}.

\begin{figure}[h!]                                           
  \centering                                                         
  \begin{minipage}{0.4\linewidth}                                
      \centering  
      \captionof{table}{Frontier-CS Benchmark Results}                           
      \resizebox{\linewidth}{!}{
      \begin{tabular}{l|cc}
      \toprule
      \textbf{Method} & \textbf{Mean} & \textbf{Median} \\
      \cmidrule(lr){1-3}
      \textbf{\adaevolve{}} & $\mathbf{61.33}$ & $\mathbf{75.15}$ \\
      OpenEvolve & $50.75$ & $56.37$ \\
      ShinkaEvolve & $47.79$ & $46.22$ \\
      GEPA & $43.04$ & $33.68$ \\
      GPT-5 (single call) & $20.64$ &  $0.0$ \\
      \bottomrule
      \end{tabular}
      }\label{tab:frontiercs_results}
  \end{minipage}%
  \hfill
  \begin{minipage}{0.6\linewidth}
      \centering
      \includegraphics[width=\linewidth]{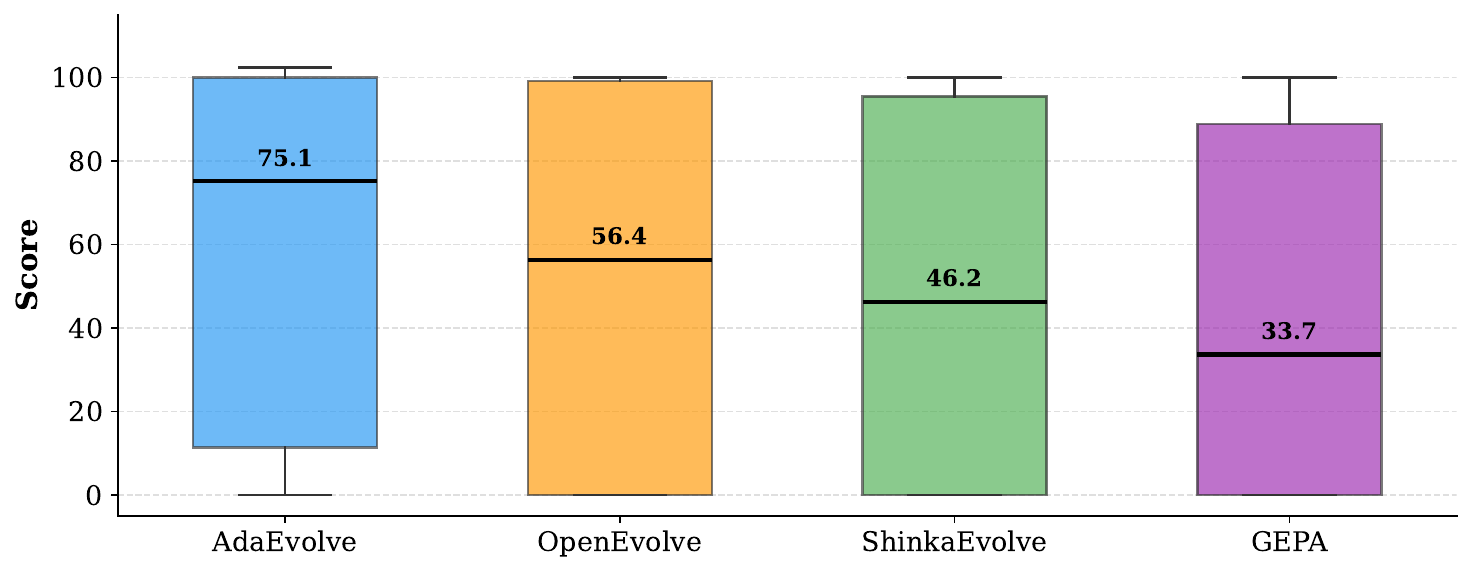}
  \end{minipage}
  \caption{Score distribution comparison across methods on the Frontier-CS benchmark.}
  \label{fig:frontiercs_comparison}
  \end{figure}


\subsection{Ablations}
\label{sec:ablations}


We evaluate the contribution of levels of adaptation in \adaevolve{}. First, we disable \textbf{Level 1: Local Adaptation} by disabling the intra-island search intensity adaptation based on the improvement signals. Instead, we use fixed 30\% exploration and 70\% exploitation probabilities as in default OpenEvolve configs. Second, we ablate different features of \textbf{Level 2: Global Adaptation}. First, we disable the bandit based island selection algorithm and replace it by simple round robin based island selection algorithm but keep spawning islands. We also test the dynamic island spawning idea by having a fixed number of 2 islands and 5 islands, respectively. Finally, we ablate \textbf{Level 3: Meta-Guidance} by disabling the generation of new solution tactics as guidance signals to the system. 

The ablations over Circle Packing and Signal Processing problems show that \adaevolve{} benefits from all different adaptive features introduced.
Meta-Guidance type of adaptation where we produce additional reflection and guidance to the LLM mutation operators during times of stagnation seem to be the most helpful feature, we its loss causes the worst results in both problems. On the other hand, bandit-based island selection seem to be more helpful for signal processing whereas local adaptation seem to be more helpful for circle packing. Fixed island experiments also indicate that having islands spawn dynamically helps allocate resources better iterations and target stagnations. 



\begin{table}[h]
\caption{Ablation results on circle packing and signal processing tasks. We report the mean and standard deviation of 3 runs.}
\centering
\begin{tabular}{l|cc}
\toprule
\textbf{Ablation Setting} & \textbf{Circle Packing} & \textbf{Signal Processing} \\
\cmidrule(lr){1-3}
\textbf{\adaevolve{}} & $\mathbf{2.6294 \pm 0.003}$ & $\mathbf{0.7178 \pm 0.019}$ \\
 w/o Local Adaptation & $2.5906 \pm 0.048$ & $0.6807 \pm 0.021$ \\
  w/o Adaptive Island Selection & $2.6180 \pm 0.005$ & $0.619 \pm 0.0541$ \\
 w/o Meta-Guidance & $2.5213 \pm 0.028$ & $0.5476 \pm 0.011$ \\
 Fixed 2 Islands & $2.6187 \pm 0.007$ & $0.5512 \pm 0.024$ \\
 Fixed 5 Islands & $2.5891 \pm 0.018$ & $0.6085 \pm 0.081$ \\
\bottomrule
\end{tabular}
\label{tab:ablation_results}
\end{table}

\subsection{Case studies: \adaevolve{} runtime adaptations}
\label{sec:case_studies}

\begin{figure}[h!]
\centering

\begin{subfigure}[t]{0.49\linewidth}
    \centering
    \includegraphics[width=0.95\linewidth]{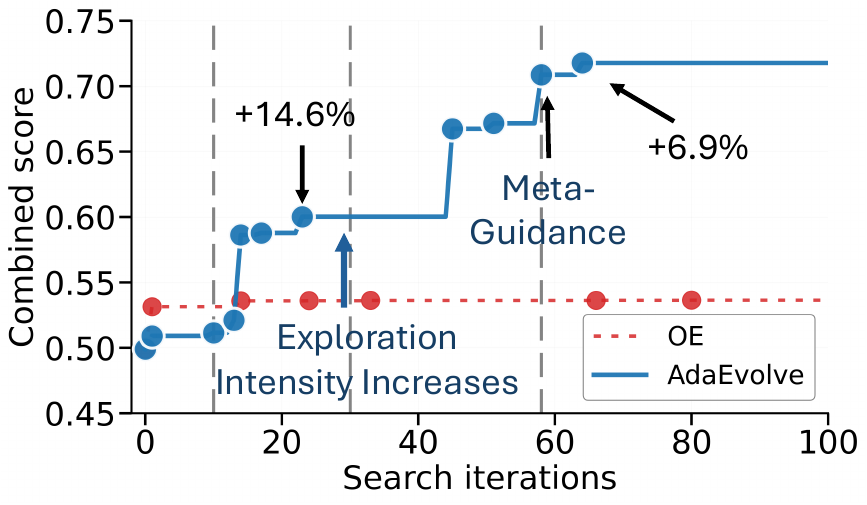}
    \caption{Signal processing run with annotations.}
    \label{fig:cs1}
\end{subfigure}
\hfill
\begin{subfigure}[t]{0.49\linewidth}
    \centering
    \includegraphics[width=0.95\linewidth]{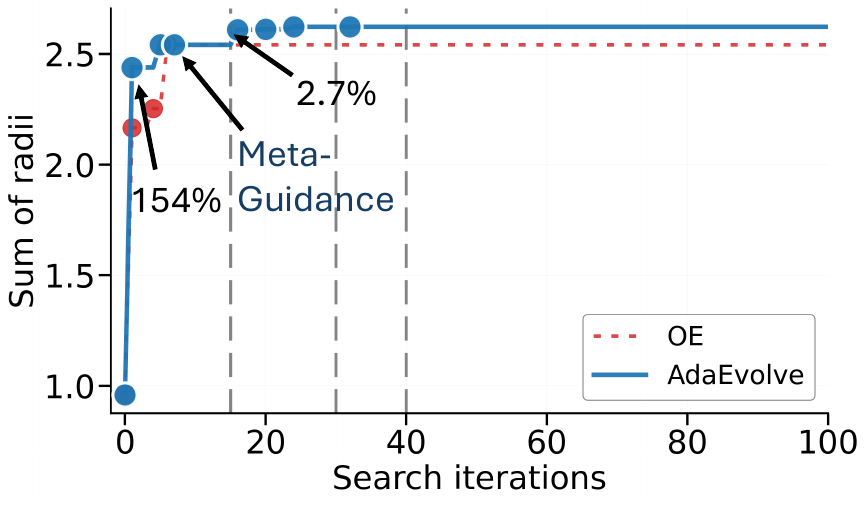}
    \caption{Circle packing run with annotations.}
    \label{fig:cs2}
\end{subfigure}

\caption{\adaevolve{} adapts search behavior across tasks.
(Left) Signal Processing: exploration transitions to refinement as improvement signals accumulate.
(Right) Circle Packing: the evolved strategy breaks stagnation and drives near-optimal layouts.}

\label{fig:case-studies}
\vspace{-4pt}
\end{figure}

We also present some detailed case studies on signal processing and circle packing tasks for \adaevolve{} using the GPT-5 model as a backbone to illustrate how different adaptive components of the algorithm benefits mitigating local search minimas


\subsubsection{Signal Processing}
\label{sec:case-study-signal}

We study the signal processing task from ~\cite{openevolve}, which requires synthesizing a causal filtering program for a noisy, non-stationary time series. Performance is measured by a combined objective capturing fidelity, smoothness, lag, and false trend changes.

Figure~\ref{fig:cs1} shows the best-so-far score of \adaevolve{} improving from $0.4990$ to $0.7177$ within 64 iterations (+43.8\%).

Early in search, the accumulated improvement signal $G_t^{(k)}$ remains near zero, inducing high search intensity and exploration-dominant sampling within islands. Parents are selected primarily for diversity, yielding only modest gains ($0.4990 \rightarrow 0.5115$ by iter~10). As improvements accumulate, $G_t^{(k)}$ increases and sampling shifts toward refinement of higher-performing parents. At iter~14, refinement using Savitzky--Golay smoothing produces a sharp improvement from $0.5210$ to $0.5862$ (+14.6\%).

As productivity becomes uneven across islands, UCB-based island selection allocates a larger fraction of iterations to the most productive island. Periodic ring migration propagates strong programs across islands, enabling continued refinement. At iter~45, refinement using an alternative low-pass operator improves performance to $0.6674$, with further gains to $0.6716$ by iter~51. Dynamic island spawning is not triggered in this run, as global improvement remains non-zero.

When numerical refinement begins to saturate, \adaevolve{} activates Meta-Guidance. The meta prompt conditions mutation on the evaluator structure and recent failure patterns, introducing alternative smoothing operators. This enables spline-based smoothing to be explored, yielding a final improvement to $0.7177$ at iter~64 (+6.9\%).

\subsubsection{Circle Packing}
\label{sec:case_study2}

Next, we study the {circle packing} task, which seeks to pack $N$ disjoint circles inside a unit square so as to maximize the sum of their radii. Figure~\ref{fig:cs2} plots the best-so-far score achieved by \adaevolve{} over optimization iterations. The score improves from $0.9598$ at initialization to $2.636$ by iter~65 (+173.4\%).

At the start of search, two islands run in parallel under adaptive exploration--exploitation, with early sampling biased toward exploration. At iter~1, random initialization discovers a dense feasible layout, improving the score from $0.9598$ to $2.4390$ (+154.2\%). Subsequent local refinement improves this layout to $2.5414$ by iter~7, after which progress stalls between iters~7 and~15.

At iter~15, global stagnation triggers Meta-Guidance, which injects an optimization-based refinement tactic. This enables continuous optimization over circle positions using SLSQP. At iter~16, exploitation applies this solution tactic to a strong layout, improving the score from $2.5414$ to $2.6095$ (+2.7\%). In contrast, runs without meta guidance remain stuck near $2.514$. Further constraint-aware refinement improves performance to $2.6121$ by iter~20, and migration propagates these improved layouts across islands.

After iter~30, sampling shifts decisively toward exploitation, concentrating search pressure on refining the best layouts rather than exploring new configurations. Local refinement of a hex-staggered configuration improves the score from $2.6121$ to $2.6228$ (+0.4\%), with additional fine-grained refinements yielding $2.6229$ at iter~32 and $2.6233$ at iter~40.

Migration at iters~45 and~60 continues to propagate the strongest layouts. Sampling remains exploitation-heavy, with local refinement focused on closing the remaining gap. Small but consistent gains are achieved at iter~55 ($2.6236$) and iter~65 ($2.636$), after which no further improvement is observed, indicating convergence.

\section{Conclusion}

In this work, we present \adaevolve{}, an adaptive evolutinary framework that only uses the fitness scores of an objective to optimize program generation for diverse algorithmic and real life systems problems. By unifying local exploration for different subpopulations of solutions, global resource routing and meta-level strategy generation under a single adaptive controller that uses the accumulated improvement signals of the past programs in the evolutionary process, our method can adapt to the non-stationary dynamics of new algorithm discovery.


Across 185 algorithm design and optimization problems, spanning 6 mathematical optimization, 7 ADRS systems tasks, and 172 algorithm design tasks from the Frontier-CS benchmark suite, we show that \adaevolve{} consistently outperforms open sourced baselines and often match or exceed the best human or AI generated solutions including proprietary models like AlphaEvolve.


\section*{Acknowledgments} 
This research has been supported by NSF (IFML) CCF-2019844 and gifts from Accenture, AMD, Anyscale, Broadcom Inc., Google, IBM, Intel, Intesa Sanpaolo, Lambda, Mibura Inc, Samsung SDS, and SAP.


\bibliographystyle{icml2025}
\bibliography{references}

\newpage
\appendix
\onecolumn
\appendix

\section{AdaEvolve Implementation Details}

\subsection{AdaEvolve Algorithm Subroutines}
\label{app:subroutines}

Here we include the subroutines mentioned in \cref{alg:adaevolve}, which are \cref{alg:paradigm}, \cref{alg:sample}, \cref{alg:mutate}, and \cref{alg:update} explained below.

\begin{algorithm}[h!]
\caption{\textsc{GenerateSolutionTactics}}
\label{alg:paradigm}
\begin{algorithmic}[1]
\Require
Global best program $p^*_{\text{global}}$; best score $f^*_{\text{global}}$;
list of past tactics $\mathcal{T}$ tried with scores;
problem specification $S$;
evaluator code $E$;
system message $S_{\text{tactic}}$

\State $\mathsf{prompt} \gets \textsc{ComposePrompt}(S, E, p^*_{\text{global}}, f^*_{\text{global}}, \mathcal{T})$
\State $\mathsf{response} \gets \textsc{LLM}(S_{\text{tactic}}, \mathsf{prompt})$
\State $\mathcal{T}_{\mathrm{new}} \gets \textsc{Parse}(\mathsf{response})$
\State $\mathcal{T} \gets \mathcal{T} \cup \mathcal{T}_{\mathrm{new}} $
\State \textbf{return} $\mathcal{T}_{\mathrm{new}}$
\end{algorithmic}
\end{algorithm}

\begin{algorithm}[h!]
\caption{\textsc{Sample} -- Adaptive Parent Selection}
\label{alg:sample}
\begin{algorithmic}[1]
\Require Archive $A$, intensity $I$
\State $u \sim \text{Uniform}(0,1)$
\If{$u < I$}
    \State \textcolor{gray}{// Exploration: favor diversity}
    \State parent $\gets$ sample from $A$ uniformly
    \State inspires $\gets$ most diverse programs from $A$
\Else
    \State \textcolor{gray}{// Exploitation: favor fitness}
    \State parent $\gets$ sample from top quartile of $A$ by fitness
    \State inspires $\gets$ highest-fitness programs from $A$
\EndIf
\State \textbf{return} parent, inspires
\end{algorithmic}
\end{algorithm}

\begin{algorithm}[h!]
\caption{\textsc{ContextBuilder}}
\label{alg:context_builder}
\begin{algorithmic}[1]
\Require Parent program $p$, inspiration programs $\{p_1, \ldots, p_m\}$, tactic $\mathcal{T}$)
\State prompt $\gets$ ``Improve the following program:'' + code($p$)
\State prompt $\gets$ prompt + ``Consider these alternative approaches:'' + code($p_1, \ldots, p_m$)
\If{tactic $\neq$ \textsc{None}}
    \State prompt $\gets$ prompt + ``Implement this strategy: '' + tactic
\EndIf
\State \textbf{return} prompt
\end{algorithmic}
\end{algorithm}

\begin{algorithm}[h!]
\caption{\textsc{Mutate}}
\label{alg:mutate}
\begin{algorithmic}[1]
\Require Model $M$, context $c$
\State $p'$ $\gets M(c)$
\State \textbf{return} $p'$
\end{algorithmic}
\end{algorithm}

\begin{algorithm}[h!]
\caption{\textsc{UpdateState} -- Adaptive State Updates}
\label{alg:update}
\begin{algorithmic}[1]
\Require Island index $k$, child fitness $f_{\text{child}}$
\State $r \gets 0$ \hfill \textcolor{gray}{// UCB reward (nonzero only on improvement)}
\If{$f_{\text{child}} > f_k^*$}
    \State $\delta \gets (f_{\text{child}} - f_k^*) / (|f_k^*| + \epsilon)$ \hfill \textcolor{gray}{// local norm.}
    \State $G_t^{(k)} \gets \rho \cdot G_{t-1}^{(k)} + (1-\rho) \cdot \delta^2$
    \State $r \gets (f_{\text{child}} - f_k^*) / (|f^*_{\text{global}}| + \epsilon)$ \hfill \textcolor{gray}{// global norm.}
    \State $f_k^* \gets f_{\text{child}}$
    \If{$f_{\text{child}} > f^*_{\text{global}}$}
        \State $f^*_{\text{global}} \gets f_{\text{child}}$
    \EndIf
\Else
    \State $G_t^{(k)} \gets \rho \cdot G_{t-1}^{(k)}$ \hfill \textcolor{gray}{// decays during stagnation}
\EndIf
\State $R_t^{(k)} \gets \rho \cdot R_{t-1}^{(k)} + r$; \quad $V_t^{(k)} \gets \rho \cdot V_{t-1}^{(k)} + 1$; \quad $n_k \gets n_k + 1$
\end{algorithmic}
\end{algorithm}

\newpage
\subsection{AdaEvolve Prompts}
\label{app:prompts}

We report the prompts used by \adaevolve{}.

\begin{promptblock}{Mode context — Exploitation}
--PARENT SELECTION CONTEXT
This parent was selected from the archive of top-performing programs. It has demonstrated strong performance, but there is likely still room for significant improvement.

--OPTIMIZATION GUIDANCE
- This solution works well, but don't assume it's optimal - meaningful improvements are still possible
- You may refine the existing approach OR introduce better algorithms if you identify a clear opportunity
- Consider: algorithmic improvements, better data structures, more efficient libraries, parallelization
- Optimizations like vectorization, caching, and numerical stability improvements are valuable
- If you see a fundamentally better approach, pursue it - but ensure correctness is maintained
- Think critically: what assumptions does this solution make? Can they be relaxed or improved?

--PARENT METRICS
\{metrics\_str\}

Your goal: Improve upon this solution - whether through refinement or strategic redesign.
\end{promptblock}

\begin{promptblock}{Mode context — Exploration}
--PARENT SELECTION CONTEXT
This parent was selected through diversity-driven sampling to explore different regions of the solution space. It may or may not represent optimal performance.

--EXPLORATION GUIDANCE
- Consider alternative algorithmic approaches or techniques
- Experiment with different methods or approaches
- Don't be constrained by the parent's approach - it's a starting point, not a template
- Look for opportunities to try fundamentally different algorithms or novel techniques
- Balance creativity with correctness - new ideas should still produce valid solutions

--PARENT METRICS
\{metrics\_str\}

Your goal: Discover new approaches that might outperform current solutions.
\end{promptblock}





\newpage

\begin{promptblock}{Solution Tactics (injected into evolution user message)}
--BREAKTHROUGH IDEA - IMPLEMENT THIS

The search has stagnated globally. You MUST implement this breakthrough idea:

**IDEA:** \{idea\}

**HOW TO IMPLEMENT:**
\{description\}

**TARGET METRIC:** \{what\_to\_optimize\}

**CAUTIONS:** \{cautions\}

**APPROACH TYPE:** \{approach\_type\}

**CRITICAL:**
- You MUST implement the breakthrough idea
- Ensure the tactics are actually used in your implementation (not just mentioned in comments)
- Correctness is essential - your implementation must be correct and functional
- Verify output format matches evaluator requirements
- Make purposeful changes that implement the idea
- Test your implementation logic carefully
\end{promptblock}

\begin{promptblock}{Tactics generator — system message ($S_{\text{tactics}}$)}
You are an expert algorithm researcher proposing \emph{breakthrough} (high-leverage) ideas.
First, deeply analyze the evaluator code to infer: the true objective, constraints/validations,
failure modes, and what is \emph{actually} rewarded or penalized.
Then analyze the current best program to identify the algorithmic approach, why it works,
and the specific bottlenecks preventing further improvement.

Propose only ideas that are correct, implementable under the available libraries, and likely to improve the score.
Each idea must be \textbf{fundamentally different} from previously tried failures, and must explicitly target a
distinct weakness or metric revealed by the evaluator. Prefer simple, robust mechanisms over complex pipelines.
Avoid suggesting vague directions: name concrete techniques, functions, and key parameters.

\textbf{Stage 1 — Understand the evaluation signal.}
Carefully read the evaluator code to determine what is actually being optimized:
the primary objective, any secondary metrics, constraints, validation rules,
and sources of penalties or failure. Identify which quantities the program can influence
and which are fixed by the evaluator.

\textbf{Stage 2 — Analyze the current best program.}
Study the current best program before proposing new ideas.
Identify the algorithmic approach it uses, why it achieves the current best score,
and what structural or algorithmic limitations are preventing further improvement.
Focus on concrete bottlenecks rather than superficial weaknesses.

\textbf{Stage 3 — Account for past attempts.}
Review previously tried tactics and their outcomes.
Do \emph{not} repeat failed approaches or close variants.
If an approach failed, reason about the underlying cause
(e.g., mismatch to problem structure, constraint violations, instability)
and avoid that class of ideas unless the root cause is explicitly addressed.

\textbf{Stage 4 — Propose diverse tactics.}
Generate multiple ideas that are \textbf{fundamentally different} from one another.
Each idea must be correct with respect to the evaluator, implementable with available libraries,
and explicitly target a distinct weakness or metric identified in earlier stages.
Each proposal should include a clear explanation of \emph{why} it is likely to improve the score.

\textbf{Stage 5 — Make ideas concrete and robust.}
Be specific and actionable: name exact libraries, functions, methods, and key parameters.
Prefer simple, robust mechanisms over complex multi-stage pipelines.
Anticipate edge cases, numerical issues, and constraint handling.

\textbf{Stage 6 — Sanity-check before answering.}
Before finalizing, verify that each idea is feasible, non-redundant with past failures,
sufficiently diverse, and aligned with what the evaluator truly rewards.
\end{promptblock}


\newpage

\subsection{Solution Tactics Generates Examples}

We present examples in Table~\ref{tab:param} for tactics generated for different use cases.
\begin{table}[h]
\centering
\small
\setlength{\tabcolsep}{4.5pt}
\renewcommand{\arraystretch}{1.12}

\caption{\textbf{Example tactics  produced by the tactic generator across domains.}
Each tactic pairs a high-level solution strategy with a representative computational approach.}
\label{tab:paradigm-examples}

\begin{tabular}{p{3.5cm} p{5.4cm} p{3.1cm}}
\toprule
\textbf{Use case} & \textbf{Meta-Guidance Solution Tactics} & \textbf{Approach type} \\
\midrule

\multirow{5}{*}{\textit{Math / equation systems}}
& Solve nonlinear systems using a trust-region root finder & \texttt{scipy.optimize.root} \\
& Exploit structure by directly solving linear systems & \texttt{scipy.linalg.solve} \\
& Isolate scalar roots with bracketed search & \texttt{scipy.optimize.brentq} \\
& Iterate toward equilibrium with damped fixed-point updates & Fixed-point iteration \\
& Generate symbolic expressions before numerical evaluation & \texttt{sympy.lambdify} + solver \\

\midrule

\multirow{5}{*}{\textit{Continuous optimization}}
& Perform constrained minimization under geometric constraints & \texttt{scipy.optimize.minimize} (SLSQP) \\
& Escape poor basins via multi-start global search & Multi-start + local refine \\
& Initialize layouts using Voronoi structure, then refine locally & Voronoi + optimizer \\
& Enforce feasibility through convex-hull constraints & \texttt{scipy.spatial.ConvexHull} \\
& Optimize box-constrained parameters with quasi-Newton updates & L-BFGS-B \\

\midrule

\multirow{5}{*}{\textit{Combinatorial / assignment}}
& Construct solutions greedily using score-ordered candidates & Greedy heuristic \\
& Compute minimum-cost matchings exactly & Linear sum assignment \\
& Improve configurations through swap-based neighborhoods & Local search \\
& Relax discrete constraints into a linear program & \texttt{scipy.optimize.linprog} \\
& Prioritize decisions with a score-driven heap & Priority-queue greedy \\

\midrule

\multirow{5}{*}{\textit{Signal / filtering}}
& Suppress impulsive noise with median filtering & \texttt{scipy.signal.medfilt} \\
& Preserve trends via polynomial smoothing & Savitzky--Golay filter \\
& Balance noise and fidelity with adaptive Wiener filtering & Wiener filter \\
& Aggregate robust statistics over sliding windows & Percentile filter \\
& Apply learned structure through custom convolution kernels & \texttt{scipy.ndimage.convolve} \\

\bottomrule
\end{tabular}
\label{tab:param}
\end{table}


\subsection{Code Structure}
\label{app:code}

\begin{archblock}{AdaEvolve architecture: main components}

\textbf{Entry and orchestration.}
The \emph{MainController} initializes the search by loading configuration,
instantiating the AdaEvolve database and manager, inserting the initial program,
and invoking a single asynchronous evolution loop. The controller does not
directly interact with islands, archives, or adaptation logic.

\vspace{0.6em}
\textbf{Evolution manager.}
The \emph{AdaEvolveManager} owns the evolution loop.
At each iteration, it (i) requests a parent and inspirations from the database,
(ii) constructs a mode-aware prompt (optionally augmented with sibling, retry,
or tactics context), (iii) generates a child program via the LLM,
(iv) evaluates the child through the evaluator, and (v) returns results to the database.
The manager never manipulates archives or adaptation state directly.

\vspace{0.6em}
\textbf{Database as system state.}
The \emph{AdaEvolveDatabase} is the sole owner of global search state.
It encapsulates islands, archives, adaptive signals, migration, spawning,
and tactics tracking. All sampling, insertion, and iteration-finalization
logic is centralized in the database.

\vspace{0.6em}
\textbf{Islands and archives.}
Islands are logical indices.
Each island maintains its population via a per-island archive.
Archives implement parent sampling, inspiration sampling, elite replacement,
and genealogy tracking, but do not perform evaluation or adaptation.

\vspace{0.6em}
\textbf{Adaptive control.}
Per-island adaptive state accumulates improvement signals and produces a
search intensity that determines exploration, exploitation, or balanced sampling.
A global UCB-based adapter selects the next island at the end of each iteration.
Adaptation state is updated only when programs are added to the database.

\vspace{0.6em}
\textbf{Evaluator.}
The evaluator is owned by the manager and invoked exactly once per generated child.
It executes the candidate program, returns metric values, and optionally emits
artifacts. The database never calls the evaluator.

\vspace{0.6em}
\textbf{Tactic mechanism.}
A tactics tracker monitors global stagnation signals maintained by the database.
When triggered, the manager injects a tactics instruction into subsequent prompts.
Tactics usage and lifecycle are tracked centrally and updated only on successful additions.

\vspace{0.6em}
\textbf{Iteration boundary.}
At the end of each iteration, the manager delegates to the database to finalize
state updates, including adaptive updates, island selection, migration,
dynamic island spawning, and checkpointing.

\end{archblock}

We outline the overall code organization and the interaction between
the controller, islands, archives, and evaluators.

\section{Benchmark Details}
\label{app:adrs}

We describe the tasks used in evaluation, including objective
definitions, evaluation costs, and sources of noise.

\subsection{ADRS}
ADRS benchmarks~\cite{cheng2025barbarians} (Table~\ref{tab:adrs_systems}) comprise real-world systems optimization tasks with discrete design choices, noisy evaluators, and heterogeneous objectives, making them representative and challenging testbeds for adaptive search.

\begin{table}[h!]
\centering
\small
\setlength{\tabcolsep}{6pt}
\renewcommand{\arraystretch}{1.25}
\caption{\textbf{ADRS systems benchmarks.}
Each task specifies a concrete systems optimization objective.}
\label{tab:adrs_systems}
\makebox[\textwidth][c]{
\begin{tabular}{p{3.6cm} p{4.8cm} p{5.8cm}}
\toprule
\textbf{Task} & \textbf{Objective} & \textbf{Problem description} \\
\midrule

\textbf{Telemetry Repair}
& Repair buggy network telemetry pipelines
& Automatically fix data processing logic to improve correctness and confidence calibration. \\

\textbf{Cloudcast}
& Optimize multi-cloud data transfer cost
& Select routing and placement strategies to minimize monetary transfer cost across providers. \\

\textbf{Expert Parallelism Load Balancer (EPLB)}
& Balance expert-parallel load across GPUs
& Reduce stragglers by redistributing expert assignments while preserving throughput. \\

\textbf{Model Placement (Prism)}
& Optimize model-to-GPU placement cost
& Assign model components to GPUs to minimize global execution cost under capacity constraints. \\

\textbf{LLM-SQL}
& Reorder tabular data for prefix efficiency
& Improve prefix-cache hit rates by reordering rows or columns without altering semantics. \\

\textbf{Transaction Scheduling (TXN)}
& Minimize makespan in transaction execution
& Schedule dependent transactions to reduce overall completion time. \\

\textbf{Datacenter TCP Congestion Control (NS3)}
& Maximize throughput and minimize queue latency
& Tune congestion control behavior under realistic network simulation workloads. \\

\bottomrule
\end{tabular}
}
\end{table}

\subsection{AlphaEvolve Math}
The AlpaEvolve~\cite{novikov2025alphaevolve} mathematical problems (Table~\ref{tab:adaevolve_problem_defs}) consist of classical combinatorial optimization problems with known formulations, providing controlled testbeds for evaluating search efficiency and convergence behavior.

\begin{table}[h!]
\centering
\small
\setlength{\tabcolsep}{6pt}
\renewcommand{\arraystretch}{1.25}
\caption{\textbf{Mathematical optimization benchmarks used in AdaEvolve.}
Each task is defined exactly as in Appendix~B of the AlphaEvolve paper.}
\label{tab:adaevolve_problem_defs}

\begin{tabular}{p{5.2cm} p{8.6cm}}
\toprule
\textbf{Problem} & \textbf{Objective (exact task definition)} \\
\midrule

\textbf{Circle Packing (Square)} \newline $N{=}26$
&
Pack $N$ disjoint circles inside a \emph{unit square} so as to
\textbf{maximize the sum of their radii}. \\

\addlinespace[0.3em]

\textbf{Circle Packing (Rectangle)} \newline $N{=}21$
&
Pack $N$ disjoint circles inside a \emph{rectangle of perimeter $4$} so as to
\textbf{maximize the sum of their radii}. \\

\addlinespace[0.3em]

\textbf{Heilbronn Problem (Triangle)} \newline $N{=}11$
&
Place $N$ points on or inside a \emph{triangle of unit area} so that
the \textbf{minimum area of any triangle formed by the points is maximized}. \\

\addlinespace[0.3em]

\textbf{Heilbronn Problem (Convex Region)} \newline $N{=}13$
&
Place $N$ points on or inside a \emph{convex region of unit area} so that
the \textbf{minimum area of any triangle formed by the points is maximized}. \\

\addlinespace[0.3em]

\textbf{MinMax Distance}  \newline $N{=}3$
&
For given $N$ and dimension $d$, find $N$ points in $\mathbb{R}^d$ that
\textbf{maximize the ratio between the minimum and maximum pairwise distances}. \\

\addlinespace[0.3em]

\textbf{Signal Processing}
&
Optimize a continuous signal processing objective under noisy evaluation,
where candidate programs transform an input signal and are scored by an
external evaluator. \\

\bottomrule
\end{tabular}
\end{table}

\newpage
\section{AdaEvolve: Additional Results}
\label{sec:arc_agi}

We evaluate AdaEvolve on an additional benchmark family to assess generalization behavior beyond optimization-centric tasks.

\textbf{ARC-AGI-2 Tasks}~\cite{chollet2025arc} evaluate abstract and compositional reasoning across 120 benchmark instances. 
Experiments follow the evaluation protocol described in Section~4.1. 
For ARC-AGI-2, OpenEvolve (OE) and \adaevolve{} operate under a matched inference budget, using 30 LLM calls per task for solution evolution on the training split, with final evaluation performed on the test split.

Unlike prior optimization benchmarks, ARC-AGI-2 is not explicitly designed for evolutionary search. 
Moreover, standard ARC evaluation assumes a strict train–test separation, whereas evolutionary frameworks typically perform adaptation at test time. 
We therefore treat this experiment as an exploratory analysis of cross-domain robustness rather than a direct optimization comparison. 
Extending evolutionary evaluation protocols to such reasoning benchmarks remains an important direction for future work.

\begin{table*}[h]
\centering
\small
\caption{AdaEvolve performance on ARC-AGI-2 benchmarks. Values denote final accuracy under a matched inference budget.}
\label{tab:ada-additional-results}

\begin{tabular}{l cccc}
\toprule
Backbone & OpenEvolve & AdaEvolve \\
\midrule
GPT-5 & 42\% & 49\% \\
Gemini-3-Pro & 44\% & 50\% \\
\bottomrule
\end{tabular}
\end{table*}

\vspace{0.5em}
\noindent
These results suggest that AdaEvolve maintains performance gains even on reasoning-oriented tasks, indicating robustness beyond traditional optimization settings.

\newpage

\end{document}